\pdfoutput=1

\documentclass[11pt]{article}

\usepackage[utf8]{inputenc}
\usepackage[T1]{fontenc}
\usepackage{xspace}
\usepackage{amsmath}
\usepackage{amsfonts}
\usepackage{hyperref}
\usepackage{url}
\usepackage{booktabs}
\usepackage{multirow}
\usepackage{makecell}
\usepackage{caption}
\usepackage{bbm}
\usepackage{graphicx}
\usepackage{balance}
\usepackage{mathtools}
\usepackage{color}
\usepackage{marvosym}
\usepackage{ifthen}
\usepackage{textcomp}
\usepackage{enumitem}
\usepackage{verbatim}
\usepackage{algorithm}
\usepackage{numprint}
\usepackage{balance}

\usepackage{amsthm}
\theoremstyle{plain}

\newcommand{\chatoDisplayMode}[1]{#1}

\definecolor{MyRed}{rgb}{0.6,0.0,0.0}
\definecolor{MyBlack}{rgb}{0.1,0.1,0.1}
\newcommand{\inred}[1]{{\color{MyRed}\sf\textbf{\textsc{#1}}}}
\newcommand{\frameit}[2]{
  \begin{center}
  {\color{MyRed}
  \framebox[.9\columnwidth][l]{
    \begin{minipage}{.85\columnwidth}
    \inred{#1}: {\sf\color{MyBlack}#2}
    \end{minipage}
  }\\
  }
  \end{center}
}

\newcommand{\note}[2][]{\chatoDisplayMode{\def\@tmpsig{#1}\frameit{{\Pointinghand} Note}{#2\ifx \@tmpsig \@empty \else \mbox{ --\em #1}\fi}}}
\newcommand{\todo}[2][]{\chatoDisplayMode{\def\@tmpsig{#1}\frameit{{\Writinghand} To-do}{#2\ifx \@tmpsig \@empty \else \mbox{ --\em #1}\fi}}}

\newcommand{\abbrevStyle}[1]{#1}

\usepackage{amssymb}
\usepackage{pifont}

\newcommand{\ie}{\abbrevStyle{i.e.}\xspace}
\newcommand{\eg}{\abbrevStyle{e.g.}\xspace}
\newcommand{\cf}{\abbrevStyle{cf.}\xspace}

\newcommand{\Eqnref}[1]{Eq.~\ref{#1}}

\newcommand{\Tabref}[1]{Table~\ref{#1}}
\newcommand{\Figref}[1]{Fig.~\ref{#1}}

\newcommand{\Appref}[1]{Appendix~\ref{#1}}

\newcommand{\xhdr}[1]{\vspace{1.7mm}\noindent{{\bf #1.}}}

\newcommand{\denselist}{ \itemsep -2pt\topsep-10pt\partopsep-10pt }

\newcommand{\textcite}[1]{\citeauthor{#1} \shortcite{#1}}

\newcommand{\hide}[1]{}

\DeclareMathOperator*{\argmax}{arg\,max}

\makeatletter
\newcommand{\iffont}[2]{\ifthenelse{\equal{\f@family}{#1}}{#2}{}}
\makeatother

  \usepackage{mathptmx}

  \DeclareSymbolFont{greek}{OML}{cmm}{m}{n}
  \DeclareMathSymbol{\alpha}{\mathalpha}{greek}{"0B}
  \DeclareMathSymbol{\beta}{\mathalpha}{greek}{"0C}
  \DeclareMathSymbol{\gamma}{\mathalpha}{greek}{"0D}
  \DeclareMathSymbol{\delta}{\mathalpha}{greek}{"0E}
  \DeclareMathSymbol{\epsilon}{\mathalpha}{greek}{"0F}
  \DeclareMathSymbol{\zeta}{\mathalpha}{greek}{"10}
  \DeclareMathSymbol{\eta}{\mathalpha}{greek}{"11}
  \DeclareMathSymbol{\theta}{\mathalpha}{greek}{"12}
  \DeclareMathSymbol{\iota}{\mathalpha}{greek}{"13}
  \DeclareMathSymbol{\kappa}{\mathalpha}{greek}{"14}
  \DeclareMathSymbol{\lambda}{\mathalpha}{greek}{"15}
  \DeclareMathSymbol{\mu}{\mathalpha}{greek}{"16}
  \DeclareMathSymbol{\nu}{\mathalpha}{greek}{"17}
  \DeclareMathSymbol{\xi}{\mathalpha}{greek}{"18}
  \DeclareMathSymbol{\pi}{\mathalpha}{greek}{"19}
  \DeclareMathSymbol{\rho}{\mathalpha}{greek}{"1A}
  \DeclareMathSymbol{\sigma}{\mathalpha}{greek}{"1B}
  \DeclareMathSymbol{\tau}{\mathalpha}{greek}{"1C}
  \DeclareMathSymbol{\upsilon}{\mathalpha}{greek}{"1D}
  \DeclareMathSymbol{\phi}{\mathalpha}{greek}{"1E}
  \DeclareMathSymbol{\chi}{\mathalpha}{greek}{"1F}
  \DeclareMathSymbol{\psi}{\mathalpha}{greek}{"20}
  \DeclareMathSymbol{\omega}{\mathalpha}{greek}{"21}
  \DeclareMathSymbol{\varepsilon}{\mathalpha}{greek}{"22}
  \DeclareMathSymbol{\vartheta}{\mathalpha}{greek}{"23}
  \DeclareMathSymbol{\varpi}{\mathalpha}{greek}{"24}
  \DeclareMathSymbol{\varrho}{\mathalpha}{greek}{"25}
  \DeclareMathSymbol{\varsigma}{\mathalpha}{greek}{"26}
  \DeclareMathSymbol{\varphi}{\mathalpha}{greek}{"27}
  \DeclareSymbolFont{otone}{OT1}{cmr}{m}{n}
  \DeclareMathSymbol{\Gamma}{\mathalpha}{otone}{0}
  \DeclareMathSymbol{\Delta}{\mathalpha}{otone}{1}
  \DeclareMathSymbol{\Theta}{\mathalpha}{otone}{2}
  \DeclareMathSymbol{\Lambda}{\mathalpha}{otone}{3}
  \DeclareMathSymbol{\Xi}{\mathalpha}{otone}{4}
  \DeclareMathSymbol{\Pi}{\mathalpha}{otone}{5}
  \DeclareMathSymbol{\Sigma}{\mathalpha}{otone}{6}
  \DeclareMathSymbol{\Upsilon}{\mathalpha}{otone}{7}
  \DeclareMathSymbol{\Phi}{\mathalpha}{otone}{8}
  \DeclareMathSymbol{\Psi}{\mathalpha}{otone}{9}
  \DeclareMathSymbol{\Omega}{\mathalpha}{otone}{10}
  \DeclareSymbolFont{syms}{OML}{cmm}{m}{it}
  \DeclareMathSymbol{\partial}{\mathord}{syms}{"40}
  \DeclareMathAlphabet{\mathbold}{OML}{cmm}{b}{it}
  \DeclareSymbolFont{largesymbols}{OMX}{cmex}{m}{n}

  \DeclareMathAlphabet{\mathcal}{OMS}{cmsy}{m}{n}

\usepackage[table]{xcolor}
\usepackage{colortbl}
\usepackage{multicol}
\usepackage[normalem]{ulem}  

\definecolor{lightblue}{rgb}{0.93,0.95,1.0}
\definecolor{lightgreen}{rgb}{0.9,1.0,0.9}
\definecolor{lightred}{rgb}{1.0,0.9,0.9}

\usepackage{natbib}
\usepackage{listings}
\usepackage{tikz}
\usepackage{subcaption}
\usepackage[many]{tcolorbox}

\DeclareSymbolFont{extraup}{U}{zavm}{m}{n}
\DeclareMathSymbol{\microsoft}{\mathalpha}{extraup}{81}
\DeclareMathSymbol{\epfl}{\mathalpha}{extraup}{83}
\DeclareMathSymbol{\psl}{\mathalpha}{extraup}{84}

\newcommand{\ourmethod}{SketchGCD}
\newcommand{\ourmethodlong}{sketch-guided constrained decoding}

\newcommand{\blackboxLLMs}{blackbox LLMs}

\newcommand{\blackbox}{blackbox}
\newcommand{\Blackbox}{Blackbox}
\newcommand{\opensourceLLMs}{open-source LLMs}

\newcommand{\sketcher}{sketcher}

\newcommand{\constrainedgenerator}{constrained decoder}

\newcommand{\stepone}{{sketching}}
\newcommand{\steptwo}{{constrained decoding}}

\newcommand{\Steptwo}{{Constrained decoding}}

\newcommand{\Gcdlong}{Grammar-constrained decoding}

\newcommand{\llama}{LLaMA}

\newcommand{\GPTfour}{GPT-4}

\newcommand{\GPTthreeFiveTurbo}{GPT-3.5-Turbo}

\usepackage{ACL2023}
\usepackage{times}
\usepackage{latexsym}
\usepackage{hyphenat}

\usepackage{microtype}

\title{Sketch-Guided Constrained Decoding for\\Boosting Blackbox Large Language Models without Logit Access}

\author{
Saibo Geng,
Berkay D\"oner,
Chris Wendler,
Martin Josifoski, 
Robert West \\
EPFL \\
{\{saibo.geng, berkay.doner, chris.wendler, martin.josifoski, robert.west\}@epfl.ch}
}

\begin{document}
\maketitle

\begin{abstract}


Constrained decoding, a technique for enforcing constraints on language model outputs, offers a way to control text generation without retraining or architectural modifications. 
Its application is, however, typically restricted to models that give users access to next-token distributions (usually via softmax logits), which poses a limitation with \blackbox{} large language models (LLMs). 
This paper introduces \textit{\ourmethodlong{}} (\ourmethod{}), a novel approach to constrained decoding for \blackbox{} LLMs, which operates without access to the logits of the \blackbox{} LLM. 
\ourmethod{} utilizes a locally hosted auxiliary model to refine the output of an unconstrained \blackbox{} LLM, effectively treating this initial output as a ``sketch'' for further elaboration.
This approach is complementary to traditional logit-based techniques and enables the application of constrained decoding in settings where full model transparency is unavailable. 
We demonstrate the efficacy of \ourmethod{} through experiments in closed information extraction and constituency parsing, showing how it enhances the utility and flexibility of \blackbox{} LLMs for complex NLP tasks.
\footnote{Code and data available at \url{https://github.com/epfl-dlab/SketchGCD}}

\end{abstract}

\section{Introduction}

\begin{figure}[t]
    \centering
    \includegraphics[width=0.99\linewidth]{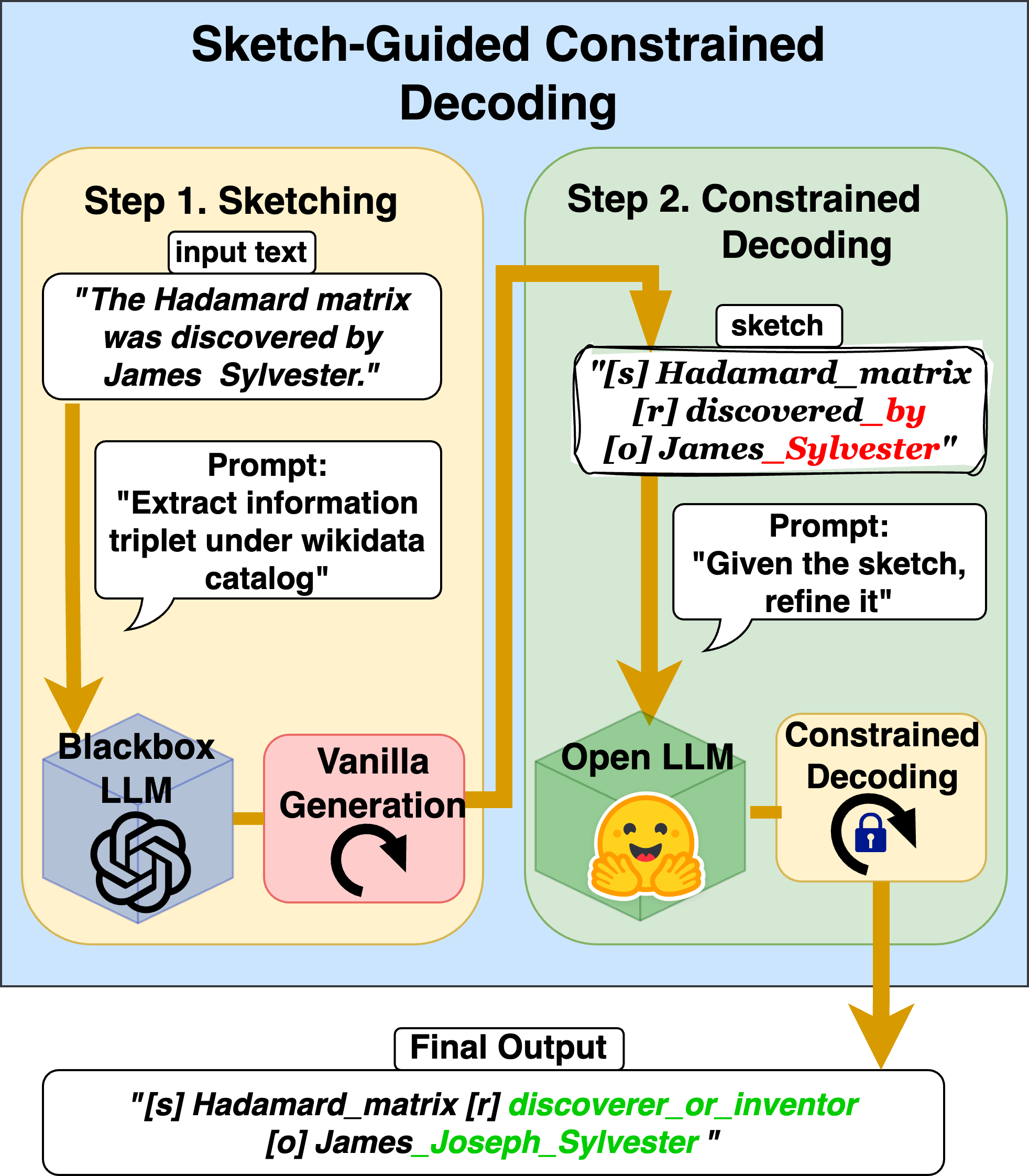}
    \caption{\textbf{Overview of \ourmethodlong{} (\ourmethod{}).} In the initial \textit{\stepone{}} phase, a \blackbox{} LLM generates a preliminary ``sketch'' answer without applying any constraints. Then, in the \textit{\steptwo{}} phase, an auxiliary model, the \constrainedgenerator{}, refines the sketch. The refined, final output respects the specified constraints by construction.}
    \label{fig:intro_figure}
\end{figure}

Large language models (LLMs) have seen a remarkable expansion in scope, being used for diverse tasks including tool interaction, SQL translation, robotic navigation and item recommendations, where adherence to specific constraints is paramount~\citep{bubeck2023sparks, schick2023toolformer, poesia2022synchromesh, shah2022lmnav, zhang2023syntax, Hua_2023_recommendation_LLM}.
Despite their versatility, LLMs often struggle with constraint adherence in few-shot scenarios, leading to outputs that violate task-specific requirements~\citep{chen2023comprehensive, agrawal_guiding_2023, huang_grounded_2023_robotics}.

Constrained decoding offers a solution, restricting model outputs to respect predefined constraints without necessitating model retraining or architectural modifications~\citep{poesia2022synchromesh, shin-etal-2021-constrained, Beurer_Kellner_2023_lmql, scholak-etal-2021-picard,geng2023grammarconstrained}.
However, existing constrained decoding methods require access to the model's logits during inference, which is not always feasible in practice (\cf{} \Appref{sec:llm-logit-access}).
Since the most powerful LLMs tend to be commercial and \blackbox{}~\citep{lee2023holistic_leaderboard}, this has restricted the application of constrained decoding methods.

\xhdr{Contributions} To overcome this restriction, we present \textit{\ourmethodlong{}} (\ourmethod{}), which bypasses the need for direct logit access.
\ourmethod{} uses a locally hosted (lightweight) open-source LLM to refine the outputs of a (heavyweight) \blackbox{} LLM to satisfy the specified constraints.
We validate our method on closed information extraction, where the constraints require generating triples grounded in a knowledge base, and constituency parsing, where the constraints require generating tree\hyp structured outputs.
Our experiments show that \ourmethod{} significantly boosts the performance of LLMs and beats previous approaches by a wide margin.

\section{Method}\label{sec:method}

\ourmethod{} splits the constrained decoding task into two distinct phases: \textit{\stepone{}} and \textit{\steptwo{}}.

During \textit{\stepone{},} a \textit{\sketcher{}}---a powerful \blackbox{} LLM denoted as $P_{\text{sk}}$---is employed. It interprets an instruction $I$ alongside a set of demonstration pairs $D=\{(x^i, y^i)\}_{i=1}^n$, producing a preliminary draft $y^*$ via unconstrained decoding:
\begin{align}
    \label{eq:sketching}
y^* \approx \argmax_{y \in S} P_{\text{sk}}(y\,|\,I,D, x),
\end{align}
where $S$ is the set of all possible sequences.

\textit{\Steptwo{}} is done by a \textit{\constrainedgenerator{},} a smaller-scale, locally hosted LLM $P_{\text{cg}}$.
Given an instruction $I_\text{cg}$, a set of input--sketch--output demonstrations $D_\text{cg}=\{(x^i, y^i, z^i)\}_{i=1}^n$, the original input $x$,
and the sketch $y^*$,
it refines $y^*$ into
%
\begin{equation}
\label{eq:constrained_generation_after_sketching}
z^* \approx \argmax_{z \in S \cap C} P_{\text{cg}}(z\,|\,I_\text{cg},D_\text{cg}, x, y^*),
\end{equation}
subject to constraints $C$.
(Optionally, $x$ and $x^i$ may be omitted, with loss of information.)
 
The \sketcher{}'s output $y^*$ is typically of high quality, encapsulating the necessary information for the \constrainedgenerator{} to produce the final sequence $z^*$ that adheres to the constraints $C$.
Given the quality of $y^*$, the \constrainedgenerator{} can be implemented using a much smaller model, as its primary task is to rewrite the sketch $y^*$ with the help of constrained decoding, thus facilitating deployment on standard consumer-grade hardware.

On the contrary, classical, direct few-shot prompting with constrained decoding would usually require a larger constrained generator $P_{\text{cg}}$ to be run locally, in order to find
\begin{equation}
\label{eq:constrained_generation}
w^* \approx \argmax_{w \in S \cap C} P_{\text{cg}}(w\,|\,I,D, x).
\end{equation}

Another basic alternative, 
unconstrained few-shot prompting \citep{brown2020language}, yields $y^*$ as the end product.

\ourmethod{} builds on the expectation that the constrained refined output $z^*$ should be at least as good as both $y^*$ (as $z^*$ respects the constraints) and $w^*$ (as $P_{\text{sk}}$ is a more powerful LLM than $P_{\text{cg}}$).

\section{Experiments}\label{sec:experimental-setup}

\begin{table*}[tb]
\centering
\resizebox{0.99\textwidth}{!}{
\setlength{\tabcolsep}{3pt}
\begin{tabular}{@{}llll|lll@{}}
\toprule
& \multicolumn{3}{c}{Wiki-NRE} & \multicolumn{3}{c}{SynthIE-text}  \\
& Precision & Recall & F1 & Precision & Recall & F1  \\
\midrule
\multicolumn{3}{l}{\emph{\textbf{Without logit access}}}\\
\hspace{4mm} \GPTfour{}                & 42.4{\scriptsize±  2.7}     &  44.9{\scriptsize±3.0}     &    43.6{\scriptsize±  2.6}   &    46.1{\scriptsize±  2.3}     &  44.4{\scriptsize±  2.2}     &    45.2{\scriptsize±  2.2}      \\
\rowcolor{lightblue}
\textbf{\hspace{9mm}  + \ourmethod{} 7B}     &    \textbf{38.7{\scriptsize±  3.2} ($\downarrow$3.7)}     &    \textbf{57.1{\scriptsize±  2.9} ($\uparrow$12.2)}     &    \textbf{46.1{\scriptsize±  2.8} ($\uparrow$2.5)}    &    \textbf{58.8{\scriptsize±  7.9} ($\uparrow$12.7)}     &  \textbf{47.3{\scriptsize±  2.3} ($\uparrow$2.9)}     &    \textbf{52.4{\scriptsize±  2.2} ($\uparrow$7.2)}    \\
\hspace{4mm} \GPTthreeFiveTurbo{}              & 27.4{\scriptsize±  2.0}     &  27.4{\scriptsize±  2.5}     &  27.4{\scriptsize±  2.5}   &    24.6{\scriptsize±  2.0}     &  23.1{\scriptsize±  1.9}     &    23.8{\scriptsize±  1.9}    \\
\rowcolor{lightblue}
\textbf{\hspace{9mm}  + \ourmethod{} 7B}   &    \textbf{31.3{\scriptsize±  3.3} ($\uparrow$3.9)}     &    \textbf{46.4{\scriptsize±  2.8} ($\uparrow$18.7)}     &    \textbf{37.4{\scriptsize±  2.8} ($\uparrow$10.0)}   &    \textbf{49.5{\scriptsize±  2.8} ($\uparrow$24.9)}     &  \textbf{41.4{\scriptsize±  2.1} ($\uparrow$18.3)}     &    \textbf{45.1{\scriptsize±  2.1} ($\uparrow$21.3)}     \\
\hspace{4mm} Claude              & 34.1{\scriptsize±  3.1}     &  28.2{\scriptsize± 2.8}     &   30.8{\scriptsize±  2.7}   &   27.0{\scriptsize±  2.0}     &  26.7{\scriptsize±  2.0}     &    26.8{\scriptsize±  2.0}     \\
\rowcolor{lightblue}
\textbf{\hspace{9mm}  + \ourmethod{} 7B}      &    \textbf{30.4{\scriptsize±  2.5} ($\downarrow$3.7)}    &    \textbf{40.6{\scriptsize±  2.8} ($\uparrow$12.4)}     &    \textbf{34.8{\scriptsize±  2.9} ($\uparrow$4.0)}   &   \textbf{51.4{\scriptsize±  2.5} ($\uparrow$24.4)}     &  \textbf{36.3{\scriptsize±  2.2} ($\uparrow$9.6)}     &    \textbf{42.5{\scriptsize±  2.2} ($\uparrow$15.7)}     \\
\hspace{4mm} Claude-instant      &  24.5{\scriptsize±  2.9}    &  18.0{\scriptsize±  2.2}     &   20.8{\scriptsize±  2.4}   &  13.0{\scriptsize±  1.7}     &  15.2{\scriptsize±  1.6}     &    14.0{\scriptsize±  1.6}   \\
\rowcolor{lightblue}
\textbf{\hspace{9mm}  + \ourmethod{} 7B} &   \textbf{44.9{\scriptsize±  3.3} ($\uparrow$20.4)}       &    \textbf{31.1{\scriptsize±  2.7} ($\uparrow$13.1)}     &    \textbf{36.7{\scriptsize±  2.5} ($\uparrow$15.9)}   &   \textbf{44.9{\scriptsize±  2.6} ($\uparrow$31.9)}     &  \textbf{31.1{\scriptsize±  2.1} ($\uparrow$15.9)}     &  \textbf{36.7{\scriptsize±  2.1} ($\uparrow$22.7)}  \\
\midrule
\multicolumn{3}{l}{\emph{\textbf{With logit access}}}\\
\hspace{4mm} \llama-2-7B     &    18.3{\scriptsize±  2.4}     &  14.0{\scriptsize±  1.8}     &    15.9{\scriptsize±  1.2}   & 12.0{\scriptsize±  1.5}     &  8.6{\scriptsize±  1.1}     &    10.0{\scriptsize±  1.3}    \\
\rowcolor{lightblue}
\textbf{\hspace{9mm}  + \ourmethod{} 7B} &   \textbf{23.6{\scriptsize±  2.7} ($\uparrow$5.3)}     &    \textbf{34.2{\scriptsize±  2.9} ($\uparrow$20.2)}     &    \textbf{28.0{\scriptsize±  2.4} ($\uparrow$12.1)}   &  \textbf{33.3{\scriptsize±  2.5} ($\uparrow$21.3)}     &  \textbf{21.0{\scriptsize±  2.0} ($\uparrow$12.4)}     &    \textbf{25.7{\scriptsize±  2.1} ($\uparrow$15.7)}    \\
\textbf{\hspace{9mm}  + CD}     &       33.6{\scriptsize±  2.7}     &    32.9{\scriptsize±  2.9}     &    32.8{\scriptsize±  2.5}   &  34.0{\scriptsize±  2.3}     &  25.9{\scriptsize±  2.0}     &    29.4{\scriptsize±  2.0}    \\
\hspace{4mm} \llama-2-13B     &      22.6{\scriptsize±  2.3}     &    23.6{\scriptsize±  2.4}     &    23.1{\scriptsize±  2.3}   &  15.7{\scriptsize±  1.6}     &  12.7{\scriptsize±  1.2}     &    14.0{\scriptsize±  1.5}    \\
\rowcolor{lightblue}
\textbf{\hspace{9mm}  + \ourmethod{} 7B} &   \textbf{28.8{\scriptsize±  2.6} ($\uparrow$6.2)}     &    \textbf{44.2{\scriptsize±  3.0} ($\uparrow$20.6)}     &    \textbf{34.9{\scriptsize±  2.5} ($\uparrow$11.8)}   &  \textbf{36.1{\scriptsize±  2.0} ($\uparrow$20.4)}     &  \textbf{25.1{\scriptsize±  1.8} ($\uparrow$12.4)}     &    \textbf{29.6{\scriptsize±  1.8} ($\uparrow$15.6)}    \\
\textbf{\hspace{9mm}  + CD}    &       35.5{\scriptsize±  2.6}     &    39.1{\scriptsize±  3.0}     &    37.2{\scriptsize±  2.5}   & 39.7{\scriptsize±  2.0}    &  32.5{\scriptsize±  1.8}     &    35.7{\scriptsize±  1.8}    \\
\hspace{4mm} \llama-2-70B     &      26.1{\scriptsize±  2.7}     &    24.5{\scriptsize±  2.3}     &    25.7{\scriptsize±  2.4}   &  32.6{\scriptsize±  2.0}     &  26.9{\scriptsize±  1.8}     &    29.4{\scriptsize±  1.8}    \\
\rowcolor{lightblue}
\textbf{\hspace{9mm}  + \ourmethod{} 7B} &   \textbf{26.9{\scriptsize±  2.7} ($\uparrow$0.8)}     &    \textbf{41.0{\scriptsize±  2.6} ($\uparrow$16.5)}     &    \textbf{32.5{\scriptsize±  2.1} ($\uparrow$6.8)}   &  \textbf{52.0{\scriptsize±  2.0} ($\uparrow$19.4)}     &  \textbf{37.6{\scriptsize±  1.8} ($\uparrow$10.7)}     &    \textbf{43.6{\scriptsize±  2.0} ($\uparrow$14.2)}    \\
\textbf{\hspace{9mm}  + CD}    &   39.9{\scriptsize±  2.6}     &    46.5{\scriptsize±  2.6}     &    42.3{\scriptsize±  2.1}   &  62.7{\scriptsize±  2.0}     &  50.3{\scriptsize±  2.0}     &    55.8{\scriptsize±  2.0}    \\
\bottomrule
\end{tabular}
}

\caption{\textbf{Results for closed information extraction,} in terms of triplet-based precision, recall, and F1-score (micro-averaged, with bootstrapped 95\% confidence intervals) on the Wiki-NRE and SynthIE-text datasets. The results compare the effectiveness of \ourmethod{} (blue rows) against two baselines:
(1) few-shot-prompted unconstrained decoding with powerful \blackbox{} LLMs (``without logit access'', white rows, \Eqnref{eq:sketching}) and
(2) few-shot-prompted constrained decoding (``CD'') with \opensourceLLMs{} (``with logit access'', \Eqnref{eq:constrained_generation}).
Four demonstrations are used in few-shot prompting.
\llama-7B serves as the constrained generator $P_\text{cg}$ for \ourmethod{}.}
\label{tab:InformationExtractionResults}
\end{table*}

%
%
%


In our experimental setup, we evaluate the efficacy of \ourmethod{} by comparing it against two established baselines:
(1) few-shot-prompted unconstrained decoding with powerful \blackbox{} LLMs (\Eqnref{eq:sketching}) and
(2) few-shot-prompted constrained decoding with \opensourceLLMs{} (\Eqnref{eq:constrained_generation}).
The \ourmethod{} method remains flexible and is agnostic to the exact implementation of constrained decoding.
Here we adopt the grammar constrained decoding framework of \citet{geng2023grammarconstrained}, but any other constraining method can be plugged in.

In our evaluation, we distinguish between sequences that are \textit{valid} (\ie, that satisfy the constraints) and those that are \textit{correct} (\ie, those that are equal to the intended output for the given input).
A {valid} output is a prerequisite for being {correct}, but it is not the sole criterion for correctness.

\subsection{Closed information extraction}\label{subsec:closed-information-extraction}

\xhdr{Task description} The goal of closed information triplet extraction (IE) is to extract a comprehensive set of facts from natural-language text.
Formally, given a knowledge base represented by a knowledge graph (KG) containing a catalog of entities $\mathcal{E}$ and a catalog of relations $\mathcal{R}$, the goal is to extract the complete set $y_\text{set} \subset \mathcal{E} \times \mathcal{R} \times \mathcal{E}$ of fact triplets expressed in a given input text $x$.
It is crucial that the entities and relations in these triplets be accurately grounded in the KG's catalog.
An example of this process can be seen in \Figref{fig:intro_figure}.
The instructions $I$ and $I_\text{cg}$ for the sketcher and constrained decoder, respectively, are listed in \Appref{app_sec:IE_instruction}.

\xhdr{Constraints} We apply the constraints in \Appref{subsec:IE_grammar}, which
restrict entities (1.5 million) and relations (857) to the Wikidata KG, and enforce the structural constraint that outputs must be formatted as sequences of entity--relation--entity triplets.

\xhdr{Datasets and evaluation metrics}
We use the Wiki-NRE~\citep{trisedya_neural_2019} and SynthIE-text~\cite{josifoski2023exploiting} datasets (details in \Appref{subsec:IE_dataset}).
Performance is measured using micro precision, recall, and F1-score.

\xhdr{Results}
We make the following observations based on \Tabref{tab:InformationExtractionResults}:
(1) The best \blackbox{} LLMs (\eg, \GPTfour{}) demonstrate strong performance even without constrained decoding, outperforming small \opensourceLLMs{} (\llama-2 7B/13B/33B) with constrained decoding.
(2) Even without requiring access to logits, \ourmethod{} still manages to enhance the performance of LLMs significantly across all models of any size.
(3) In case where logit access is available, constrained decoding is more effective than \ourmethod{}, as shown by the second half of the table.
Given these observations, we conjecture that, if logits were accessible for \blackbox{} LLMs, a further improvement in performance could be achieved with constrained decoding.
However, without logit access, \ourmethod{} provides an effective alternative.

\xhdr{Impact of \constrainedgenerator{}}
\begin{table}[tb]
\centering
\resizebox{\columnwidth}{!}{
\setlength{\tabcolsep}{7pt}
\begin{tabular}{@{}lccc|ccc@{}}
\toprule
& \multicolumn{3}{c}{Wiki-NRE} & \multicolumn{3}{c}{SynthIE-text} \\
& Prec & Recall & F1 & Prec & Recall & F1 \\
\midrule
 \GPTfour{} & 42.4 & 44.9 & 43.6 & 46.1 & 44.4 & 45.2 \\
\hspace{4mm} + \llama-2-7B & 38.7 & \textbf{57.1} & 46.1 & \textbf{58.9} & 47.3 & 52.4 \\
\hspace{4mm} + \llama-2-13B & \textbf{42.9} & 52.8 & \textbf{47.3} & 53.6 & 51.4 & 52.5 \\
\hspace{4mm} +  \llama-2-70B & 35.2 & 54.0 & 42.6 & 58.1 & \textbf{53.1} & \textbf{55.5} \\
\bottomrule
\end{tabular}
}
\caption{\textbf{Impact of constrained decoder model} (used in step 2 of \ourmethod{}) on closed information extraction. \GPTfour{} is used as the sketcher in all cases.}
\label{tab:IE_many_decoder_analysis}
\end{table}
We investigate the impact of the \constrainedgenerator{} on the performance of \ourmethod{}.
As shown in \Tabref{tab:IE_many_decoder_analysis}, given \GPTfour{} as the \sketcher{}, the choice of the \constrainedgenerator{} can affect the performance of \ourmethod{}.
Contrary to our expectations, larger \constrainedgenerator{} models do not always lead to better performance.
Our intuition is that step 2 of \ourmethod{} (constrained decoding) is relatively simple, and the additional capacity of larger \constrainedgenerator{}s does not necessarily provide an advantage.

\xhdr{Impact of beam size}
Our experiments show that using beam search is critical for the performance of both \ourmethod{} and classical constrained decoding.
As shown in \Tabref{tab:IE_beam_search_analysis}, employing beam search (even with a minimal beam size of 2) significantly improves performance over greedy decoding.
Larger beam sizes further enhance performance, allowing the model to explore a larger search space, but with a diminishing returns.

The following example illustrates the importance of beam search.
Suppose we are doing closed information extraction on the sentence \textit{``Mona Lisa is housed in the Musée du Louvre in Paris.''}
    Our entity catalog contains among other, the entities \textit{Louvre Museum} and \textit{Musée d'Orsay}.
    During unconstrained decoding, the model might generate the following output with highest probability:
    \textit{``[s] Mona Lisa [r] located in [o] Musée \textbf{du Louvre}''}.
    This output is invalid as the entity \textit{Musée du Louvre} is not in the entity catalog and should be rendered as \textit{Louvre Museum} instead.

    With constrained decoding, the non-bold part of the output remains unaltered, as it satisfies the constraints.
    However, the bold suffix \textit{``du Louvre''} is rejected by constrained decoding because \textit{Mus\'ee du Louvre} is not in the entity catalog.
    The model will be forced to sample from the allowed entity catalog only, which can lead to \textit{``Musée d'Orsay''} as the output.
    In this example, greedy constrained decoding was able to produce a valid yet incorrect output.
    On the contrary, had we used beam search, the model would have been able to consider both  \textit{Musée du Louvre} and \textit{Louvre Museum} simultaneously, and would have been able to select the correct entity, \textit{Louvre Museum}, for the output.
\begin{table}[tb]
\centering
\resizebox{0.48\textwidth}{!}{
\setlength{\tabcolsep}{7pt}
\begin{tabular}{@{}lccc|ccc@{}}
\toprule
& \multicolumn{6}{c}{Wiki-NRE}  \\
& \multicolumn{3}{c}{\llama-2-7B + CD} & \multicolumn{3}{c}{\llama-2-13B + CD} \\
& Prec & Recall & F1 & Prec & Recall & F1 \\
\midrule
\hspace{4mm} 1 beam & 29.9 & 22.6 & 25.8 & 32.7 & 32.3 & 32.5 \\
\hspace{4mm} 2 beams & 33.6 & 32.1 & 32.8 & 35.9 & \textbf{39.6} & 37.7 \\
\hspace{4mm} 4 beams & 33.7 & \textbf{32.9} & 33.3 & 36.0 & 38.5 & 37.2 \\
\hspace{4mm} 8 beams & \textbf{36.6} & 30.8 & \textbf{33.4} & \textbf{39.6} & 36.0 & \textbf{37.7} \\
\bottomrule
\end{tabular}
}
\caption{\textbf{Impact of beam size} in beam search on closed information extraction during classical constrained decoding. ``1 beam'' is equivalent to greedy decoding. }
\label{tab:IE_beam_search_analysis}
\end{table}

\subsection{Constituency parsing}
\begin{table*}[h]
\centering
\resizebox{0.99\textwidth}{!}{
\setlength{\tabcolsep}{5pt}
\begin{tabular}{@{}lllllll@{}}
\toprule
\hspace{4mm} Method & Bracket prec* & Bracket recall* & Bracket F1* & Tag accuracy* & Tag validity & Tree validity\\
\midrule
\multicolumn{3}{l}{\emph{\textbf{Without logit access}}}\\
\hspace{4mm}\GPTfour{} & 76.6{\scriptsize± 5.0} & 67.7{\scriptsize± 4.5} & 71.9{\scriptsize± 4.0} & 95.4{\scriptsize± 0.9} & 93.6{\scriptsize± 4.0} & 86.0{\scriptsize± 4.0} \\
\rowcolor{lightblue}
\textbf{\hspace{8mm} + \ourmethod{}} & \textbf{75.8{\scriptsize± 2.4} ($\downarrow$0.8)} & \textbf{67.8{\scriptsize± 2.4} ($\downarrow$0.1)} & \textbf{71.5{\scriptsize± 2.4} ($\downarrow$0.4)} & \textbf{95.3{\scriptsize± 0.8} ($\downarrow$0.1)} &  \textbf{100{\scriptsize± 0.0} ($\uparrow$6.4)} & \textbf{92.5{\scriptsize± 4.0} ($\uparrow$6.5)} \\
\hspace{4mm}\GPTthreeFiveTurbo{} & 68.2{\scriptsize± 0.7} & 55.5{\scriptsize± 1.1} & 61.2{\scriptsize± 0.6} & 93.1{\scriptsize± 0.5} & 91.7{\scriptsize± 2.4} & 76.9{\scriptsize± 5.2} \\
\rowcolor{lightblue}
\textbf{\hspace{8mm} + \ourmethod{}} & \textbf{68.7{\scriptsize± 3.2} ($\uparrow$0.5)} & \textbf{56.6{\scriptsize± 2.8} ($\uparrow$1.1)} & \textbf{62.1{\scriptsize± 2.0} ($\uparrow$0.9)} & \textbf{92.6{\scriptsize± 1.3} ($\downarrow$0.5)} &  \textbf{100{\scriptsize± 0.0} ($\uparrow$8.3)} & \textbf{81.5{\scriptsize± 4.0} ($\uparrow$4.6)} \\
\hspace{4mm}Claude 2.1 & 73.1{\scriptsize± 3.3} & 63.1{\scriptsize± 2.6} & 67.7{\scriptsize± 2.5} & 94.5{\scriptsize± 1.1} & 95.1{\scriptsize± 2.5} & 62.6{\scriptsize± 5.2} \\
\rowcolor{lightblue}
\textbf{\hspace{8mm} + \ourmethod{}} & \textbf{71.6{\scriptsize± 3.0} ($\downarrow$1.5)} & \textbf{62.9{\scriptsize± 2.5} ($\downarrow$0.2)} & \textbf{66.9{\scriptsize± 2.6} ($\downarrow$0.8)} & \textbf{93.4{\scriptsize± 1.3} ($\downarrow$1.1)} &  \textbf{100{\scriptsize± 0.0} ($\uparrow$4.9)} & \textbf{68.7{\scriptsize± 5.5} ($\uparrow$6.1)} \\
\hspace{4mm}Claude-instant 1.2 & 71.3{\scriptsize± 2.4} & 59.1{\scriptsize± 1.4} & 64.7{\scriptsize± 1.9} & 89.6{\scriptsize± 1.6} & 91.7{\scriptsize± 2.3} & 56.6{\scriptsize± 5.2} \\
\rowcolor{lightblue}
\textbf{\hspace{8mm} + \ourmethod{}} & \textbf{66.6{\scriptsize± 3.3} ($\downarrow$4.7)} & \textbf{57.4{\scriptsize± 3.1} ($\downarrow$1.7)} & \textbf{61.6{\scriptsize± 3.3} ($\downarrow$3.1)} & \textbf{87.9{\scriptsize± 2.5} ($\downarrow$1.7)} &  \textbf{100{\scriptsize± 0.0} ($\uparrow$8.3)} & \textbf{67.8{\scriptsize± 3.7} ($\uparrow$11.2)} \\
\midrule
\multicolumn{3}{l}{\emph{\textbf{With logit access}}}\\
\hspace{4mm} llama-2-7B & 23.1{\scriptsize± 4} & 10.4{\scriptsize± 3} & 14.3{\scriptsize± 4} & 14.9{\scriptsize± 3} & 93.2{\scriptsize± 3} & 32.1{\scriptsize± 5} \\
\rowcolor{lightgreen}
\hspace{8mm}  + \textbf{CD} & 28.5{\scriptsize± 6} ($\uparrow$5.4) & 16.5{\scriptsize± 3} ($\uparrow$6.1) & 20.9{\scriptsize± 5} ($\uparrow$6.6) & 13.8{\scriptsize± 2} ($\downarrow$1.1) & 100{\scriptsize± 0} ($\uparrow$6.8) & 35.1{\scriptsize± 5} ($\uparrow$3.0) \\
\hspace{4mm} llama-2-13B & 33.4{\scriptsize± 7} & 22.4{\scriptsize± 4} & 26.8{\scriptsize± 5} & 29.3{\scriptsize± 4} & 95.5{\scriptsize± 2} & 38.5{\scriptsize± 6} \\
\rowcolor{lightgreen}
\hspace{8mm} + \textbf{CD} & 33.3{\scriptsize± 6} ($\downarrow$0.1) & 21.8{\scriptsize± 5} ($\downarrow$0.6) & 26.3{\scriptsize± 5} ($\downarrow$0.5) & 34.0{\scriptsize± 4} ($\uparrow$4.7) & 100{\scriptsize± 0} ($\uparrow$4.5) & 43.4{\scriptsize± 5} ($\uparrow$4.9) \\
\hspace{4mm} llama-2-70B & 45.5{\scriptsize± 6} & 37.7{\scriptsize± 5} & 41.2{\scriptsize± 5} & 55.5{\scriptsize± 5} & 75.8{\scriptsize± 5} & 40.4{\scriptsize± 6} \\
\rowcolor{lightgreen}
\hspace{8mm}  + \textbf{CD} & 39.8{\scriptsize± 6} ($\downarrow$5.7) & 35.6{\scriptsize± 4} ($\downarrow$2.1) & 37.6{\scriptsize± 4} ($\downarrow$3.6) & 53.8{\scriptsize± 4} ($\downarrow$1.7) & 100{\scriptsize± 0} (($\uparrow$24.2)) & 47.6{\scriptsize± 5} ($\uparrow$7.2) \\
\bottomrule
\end{tabular}
}
\caption{\textbf{Results for constituency parsing,}
in terms of bracketing precision, recall, F1-score, tag accuracy, tag validity, and parse tree validity (with bootstrapped 95\% confidence intervals),
on Penn Treebank test split.
Only subset of samples whose ground-truth parse trees are shorter than 128 tokens (per \llama{} tokenizer) are considered (shortest 25\% of the full dataset).
Disclaimer: a weak method can have high precision by predicting very few valid parse trees (simple ones), and a strong method can have low precision by predicting more valid parse trees including complex ones \citep{deutsch-etal-2019-general}.
Four demonstrations are used in few-shot prompting.
LLaMA-7B serves as the constrained generator $P_\text{cg}$ for SGCD.
(* Considering only sentences with valid parse trees.)
}
\label{tab:CP_main_results}
\end{table*}

\xhdr{Task description}
Constituency parsing involves breaking down a sentence into its syntactic components to form a parse tree that represents the sentence's structure.
For instance, the sentence \textit{``I saw a fox''} corresponds to the parse tree [S [NP [PRP \textit{I}]] [VP [VBD \textit{saw}] [NP [DT \textit{a}] [NN \textit{fox}]]]].
For a visual representation of this tree, see Appendix~\ref{app_sec:CP}~\Figref{fig:example_parse_trees}.
The instructions $I$ and $I_\text{cg}$ are listed in \Appref{app_sec:CP_instruction}.

\xhdr{Constraints}
We apply the context-free grammar constraints in \Appref{subsec:CP_grammar} to ensure that brackets are balanced, and labels are consistent.

\xhdr{Dataset and evaluation metrics}
Our evaluation uses the Penn Treebank test split.
The parsing error rate of LLMs, regardless of size, is generally high, so we use only the shortest 25\% of the samples for evaluation (up to 128 tokens according to the \llama{} tokenizer).
We assess performance using bracketing recall and precision, as well as tag accuracy, as measured by the EVALB tool~\cite{evalb}.
Since these metrics are only applicable to valid parse trees, and since models typically generate valid trees only for simpler inputs, one needs to be careful while interpreting the results, as weaker model may have better scores because they only generate a small fraction of valid parse trees (simpler ones)~\cite{deutsch-etal-2019-general}.

\xhdr{Results}
The results in \Tabref{tab:CP_main_results} show that even advanced LLMs like \GPTfour{} struggle to generate valid parse trees, especially for longer sentences.
The following observations can be made:
(1) Both \ourmethod{} and classical constrained decoding significantly help the model generate more structurally valid parse trees.
(2) The other metrics mostly remain unchanged or slightly drop, as a larger validity rate means more difficult examples are included in the evaluation.
(3) The most common errors in the unconstrained setting are \textit{imbalanced brackets}, \textit{invalid tags}, and \textit{missing words,} as shown in \Tabref{tab:error_metrics_by_method}.
(4) With \ourmethod{}, the error rate for \textit{imbalanced brackets} and \textit{invalid tags} is significantly reduced, while the error rate for \textit{missing words} increases significantly.

Note that constrained decoding with a more sophisticated grammar, as described in \Appref{subsec:CP_grammar},  can achieve 100\% valid trees and 100\% valid tags (see \Tabref{tab:two_grammar_cp_results}).
However, as implementing such a grammar is non-trivial, we use a simpler context-free grammar here (see \Appref{subsec:CP_grammar}) to mimic the real-world scenario where a simpler might be preferred over a perfect grammar.

\begin{table}[h]
\centering
\resizebox{0.5\textwidth}{!}{
\setlength{\tabcolsep}{2pt}
\begin{tabular}{@{}lcccc@{}} 
\toprule
& \multicolumn{4}{c}{Error type} \\
Method  & InvalidTag & Extra & Imbal & Missing  \\
\midrule
\GPTfour{}  & 6.4\% & 0.4\% & 10.2\% & 2.3\% \\
\hspace{4mm} + \textbf{\ourmethod{}} & 0.0\% & 0.0\% & 2.6\% & 6.0\% \\
\GPTthreeFiveTurbo{}  & 8.3\% & 2.6\% & 9.4\% & 2.3\% \\
\hspace{4mm} + \textbf{\ourmethod{}} & 0.0\% & 1.5\% & 1.9\% & 16.2\% \\
Claude 2.1 & 4.9\% & 3.8\% & 3.4\% & 30.2\% \\
\hspace{4mm} + \textbf{\ourmethod{}} & 0.0\% & 3.0\% & 3.8\% & 29.8\% \\
%
\bottomrule
\end{tabular}
}
\caption{\textbf{Error analysis for constituency parsing} on the Penn Treebank dataset. \textit{InvalidTag} refers to model generating invalid tags, \textit{Extra} to model adding extra words absent from input, \textit{Imbal} to model generating imbalanced brackets, and \textit{Missing} to model dropping words from input.}
\label{tab:error_metrics_by_method}
\end{table}

\section{Related work}\label{sec:related_work}

%
%

\xhdr{Constrained decoding}
\citet{deutsch-etal-2019-general} introduced a general constrained decoding framework for text generation based on automata.
\citet{scholak-etal-2021-picard, poesia2022synchromesh, geng2023grammarconstrained} implemented incremental parsing for domain-specific tasks such as SQL generation.
\citet{Beurer_Kellner_2023_lmql, poesia2023certified} have proposed iterative approaches to constrained decoding using \blackbox{} LLM APIs, albeit with potential limitations such as excessive API calls (thus increasing monetary cost), as detailed in \Appref{app_sec:logit_bias_based_method}.

\xhdr{Collaborative generation}
\citet{vernikos2023small} and \citet{welleck2022generating} explored training smaller language models to refine the outputs from larger models for enhanced quality.
The skeleton-of-thought method~\citep{ning_skeleton--thought_2023}  generates an initial output skeleton and then concurrently develops each segment.
Grammar prompting~\citep{bailin_wang2023grammar} creates a meta-grammar to guide the output of LLMs in producing valid results.

\section{Conclusion}\label{sec:conclusion}
So far, constrained decoding has been limited to open-source models that provide access to their logits during generation.
Overcoming this limitation, we propose \ourmethodlong{} (\ourmethod{}), a simple method for constrained decoding with \blackbox{} LLMs that does not require access to next-token logits during generation.
By using separate sketching and refinement phases, \ourmethod{} allows to benefit from the power of \blackbox{} LLMs while still enforcing constraints.
Our work is complementary to existing methods for constrained decoding and can be used in conjunction with them.
%
Despite its simplicity, \ourmethod{} achieves strong performance on tasks exhibiting strong structural constraints, outperforming unconstrained generation by a large margin.


\section{Limitations}\label{sec:limitations}

The limitations of our method include the following.
First, \ourmethod{} adds an overhead as it requires a \constrainedgenerator{} to refine the sketches after the sketching phase. 
Second, as LLMs keep getting better, the benefits of \ourmethod{} might diminish on some tasks as the unconstrained model's performance improves.
Third, just as classical constrained decoding, \ourmethod{} can only enforce constraints at the structure level or the syntactic level, but not at the semantic level. The model can still generate semantically incorrect outputs.
However, in many real-world applications, we have observed semantic errors to be less common than structural errors.

\section*{Acknowledgments}

We thank Grant Slatton for insightful discussions during the ideation phase.
West's lab is partly supported by grants from
Swiss National Science Foundation (200021\_185043, TMSGI2\_211379),
Swiss Data Science Center (P22\_08),
H2020 (952215),
Microsoft Swiss Joint Research Center,
and Google,
and by generous gifts from Meta, Google, and Microsoft.



\bibliography{custom}

\clearpage

\appendix

\section{\Blackbox{} LLM logit access}\label{sec:llm-logit-access}

\begin{table}[htbp]
\centering
\resizebox{\linewidth}{!}{
\begin{tabular}{lrrr}
\toprule
\textbf{Model}  & \textbf{Logit bias} & \textbf{Token probs} & \textbf{MMLU} \\
\midrule
\GPTfour{}-0614             & Yes & Top 5 & 86.4 \\
\GPTthreeFiveTurbo{}-0614             & Yes & Top 5 & 70.0 \\
Claude-2.1             & No & No & 78.5 \\
Claude-instant      & No & No & 73.4 \\
PaLM-2-text-bison   & Yes & Top 5 & 78.3 \\
\bottomrule
\end{tabular}
}
\caption{The \blackbox{} LLMs we use in our experiments and the access they provide to the logit distribution. MMLU is the mainstream metric for LLM benchmarking.
}
\label{tab:llm-logit-access}
\end{table}

\textit{Logit bias} indicates whether the model's API allows user to pass in a logit bias vector to steer the decoding process, \ie, \textit{write access} to the logit distribution.
\textit{Token probs} indicates whether the model's API allows user to access the model's next token probability distribution, \ie, \textit{read access} to the logit distribution.
MMLU~\citep{hendrycks2021measuring_MMLU} is the mainstream metric for LLM benchmarking.

\section{Grammar constrained decoding}\label{sec:grammar-constrained-decoding}

\Gcdlong{} takes a formal grammar $G$ as input and ensures that the output string $w$ is a valid \textit{sentence} in the formal language $L(G)$ defined by the grammar $G$.
This process is achieved through the integration of two key components: a \textit{grammar completion engine}~\citep{poesia2022synchromesh} and a \textit{sampling method}, \eg greedy search, nucleus sampling, etc.
The grammar completion engine is used to ensure the \textit{grammaticality} of the output string, while the LLM is used to ensure the \textit{plausibility} of the output string.

We use \textit{Grammatical Framework}'s runtime powered completion engine~\cite{ranta-2019-grammatical} with constrained beam search as the sampling method.

\section{Logit bias-based iterative decoding}\label{app_sec:logit_bias_based_method}

Most \blackbox{} LLM APIs do not provide complete access to the model's next token probability distribution at each decoding step.
Nonetheless, many allow users to input a \textbf{logit bias} parameter to influence the decoding process, \ie, granting users \textit{write access} but not \textit{read access} to the model's logits at each decoding step.
This parameter accepts a vector of logits that is added to the logits of the next token probability distribution at each decoding step.
By using the logit bias parameter, users can direct the decoding process, effectively masking the logits of invalid tokens.
This approach is particularly effective for static constraints, such as lexical constraints~\citep{hokamp-liu-2017-lexically}, where the constraints remain constant throughout the decoding.

However, the logit bias parameter is a static array and does not change during the decoding process.
This makes it challenging to apply dynamic constraints, which change as decoding progresses, such as constraints involving membership in formal languages~\citep{deutsch-etal-2019-general, poesia2022synchromesh, geng2023grammarconstrained}.

A straightforward but costly solution for dynamic constraints is to iteratively invoke the \blackboxLLMs{} API with updated logit bias vectors at each decoding step~\citep{Beurer_Kellner_2023_lmql, poesia2023certified, agrawal_guiding_2023, choi_kcts_2023}.
However, this approach is \textbf{prohibitively expensive}.
Each API call generates only a single token, and the cost is calculated based on both the input and output tokens\footnote{See \url{https://openai.com/pricing} for details.}.
The expense of iteratively calling the \blackboxLLMs{} API with new context and prefix at each step scales quadratically, being $O(n^2)$ where $n$ is the length of the output sequence.
Although methods like those proposed by \citet{Beurer_Kellner_2023_lmql} and \citet{poesia2023certified} use speculation to reduce the number of API calls, the costs can remain high, especially when the constraints are complex.

\section{Task 1. closed information extraction}\label{app_sec:IE}

In this section, we provide more details about the closed information extraction task.

\subsection{Task instruction}\label{app_sec:IE_instruction}

We provide the instruction for the IE task in Figure~\ref{fig:IE_instructions}.
The few-shot demonstrations are rather long and thus we do not include them here.
The full prompt is available in our code repository.

\begin{figure}[ht]
\centering
\begin{subfigure}[b]{\columnwidth}
\centering
\begin{tikzpicture}
\node[draw, rectangle, rounded corners, fill=blue!20, inner sep=10pt, text width=0.9\columnwidth, align=center]
    at (0,0) {Extract the subject-relation-object triples in fully-expanded format from texts below. The subjects and objects are entities in Wikidata, and the relations are Wikidata properties. Here are a few examples.};
\end{tikzpicture}
\caption{Instruction for \sketcher{}}
\label{fig:IE_instruction_sketcher}
\end{subfigure}

\begin{subfigure}[b]{\columnwidth}
\centering
\begin{tikzpicture}
\node[draw, rectangle, rounded corners, fill=red!20, inner sep=10pt, text width=0.9\columnwidth, align=center]
    at (0,0) {In this task, you will be provided with texts along with draft annotations that represent extracted information triples in the form of subject-relation-object. Your role is to refine these triples to ensure completeness and accuracy. Here are a few examples.};
\end{tikzpicture}
\caption{Instruction for \constrainedgenerator{}}
\label{fig:IE_instruction_constrained_generator}
\end{subfigure}

\caption{Instructions for parsing tasks.}
\label{fig:IE_instructions}
\end{figure}

\subsection{Grammar}\label{subsec:IE_grammar}

The grammar is defined as follows, where \( V \) represents the set of variables, \( \Sigma \) the set of terminal symbols, and \( P \) the set of production rules:

\begin{align*}
&V = \{S, T, A, B, C, E, R\},  \Sigma = \{\texttt{tokens}\} \\
&P = \{S \rightarrow [S T | \epsilon], T \rightarrow [A B C \texttt{[e]}] \\
&A \rightarrow [\texttt{[s]}\; E], E \rightarrow (\texttt{entity1} | \texttt{entity2} | ...), \\
&B \rightarrow [\texttt{[r]}\; R], R \rightarrow (\texttt{rel1} | \texttt{rel2} | ...) \\
&C \rightarrow [\texttt{[o]}\; E], \epsilon \rightarrow \texttt{</s>}\}
\end{align*}

The outputs are structured as a sequence of triplets, where each triplet is separated by a special marker \texttt{[e]}.
Every triplet consists of a subject, a relation, and an object.
These elements are each preceded by a special marker: \texttt{[s]} for the subject, \texttt{[r]} for the relation, and \texttt{[o]} for the object, respectively.
The subject and object are pre-defined Wikidata entities, and the relation is a pre-defined Wikidata property.
This grammar is classified as context-free, more specifically, as a regular grammar.

\subsection{IE datasets}\label{subsec:IE_dataset}
%
%
%

The original SynthIE-text and Wiki-NRE datasets comprise 50,000 and 30,000 samples, respectively.
To minimize the evaluation cost on Large Language Models (LLMs), we use a smaller subset consisting of 1,000 samples from each dataset.

\begin{figure}[ht]
    \centering
    \begin{subfigure}[b]{0.45\textwidth}
    \includegraphics[width=\textwidth]{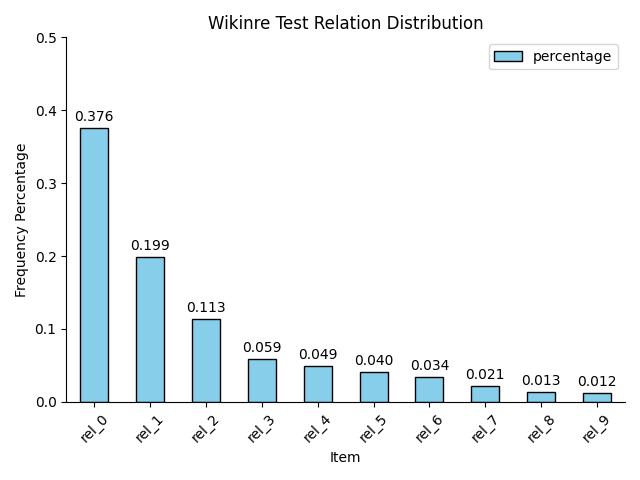}
    \caption{Original relation distribution in WikiNRE test set}
    \label{fig:IE_relation_distribution}
    \end{subfigure}
    \begin{subfigure}[b]{0.45\textwidth}
    \includegraphics[width=\textwidth]{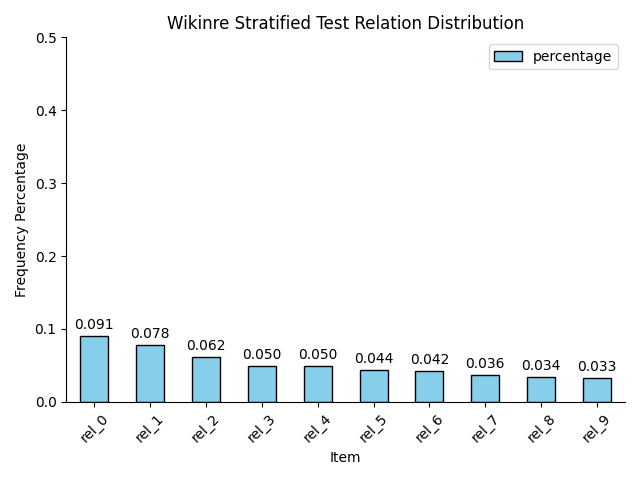}
    \caption{Stratified relation distribution in WikiNRE test set}
    \label{fig:IE_relation_distribution_wn18}
    \end{subfigure}
    \caption{Relation distribution in WikiNRE before and after stratification.}
    \label{fig:relation_distribution}
\end{figure}

As noted by \citet{josifoski2023exploiting}, the Wiki-NRE dataset displays a significant skew in its relations distribution: the top 10 relations constitute 92\% of the triplets, with the top 3 alone accounting for 69\%.
To ensure our test set accurately reflects the overall dataset, we have downscaled it to 1,000 samples to balance the distribution of relations, as shown in \Figref{fig:relation_distribution}

The SynthIE-text dataset, synthesized by reverse prompting \textbf{Text-Davinci-003} with triplets from Wikidata, stands out due to its substantial size, diverse content, and high-quality human ratings, as highlighted in~\cite{josifoski2023exploiting}.
This contrasts with prior datasets such as REBEL \cite{huguet-cabot-navigli-2021-rebel-relation}, whose annotation quality is low~\citep{josifoski-etal-2022-genie}.
However, a potential minor bias may exist towards \GPTfour{} and \GPTthreeFiveTurbo{}, as SynthIE-text was generated from a model in their family, \textbf{Text-Davinci-003}.
Despite this, we maintain that this does not compromise the validity of our method, given that our primary focus is on the comparative performance with and without the application of \ourmethod{}.

\subsection{Discussion of \GPTfour{} on cIE}\label{subsec:IE_gpt4_analysis}

An intriguing finding is that \ourmethod{}'s performance on the SynthIE-text dataset using \GPTfour{} (F1=45.6) is marginally lower than that achieved with few-shot prompting alone, without constrained decoding (F1=45.8).
Our analysis suggests that the \constrainedgenerator{} occasionally struggles to adhere to the \sketcher{}'s outline, resulting in fewer triplets than expected in the output.
This observation is consistent with \citet{bailin_wang2023grammar}'s findings, where constrained decoding was noted to reduce the diversity of the generated samples.

More critically, since SynthIE-text is synthetically generated by reverse prompting \textbf{Text-Davinci-003} with triplets from Wikidata, its text doesn't exhibit the naturalness characteristic of the Wiki-NRE dataset.
For instance, sentences in SynthIE-text often resemble direct copies with slight alterations from the original entity and relation names.
This tendency facilitates the LLMs' task of grounding entities and relations in the Knowledge Graph (KG), thereby diminishing the necessity for constrained decoding.

However, in real-world scenarios, text is typically more intricate, and grounding entities and relations in the KG is not as straightforward.
Despite this, the overall performance enhancement provided by \ourmethod{} across various models remains noteworthy, averaging gains of up to 10.7\% and 8.1\% on Wiki-NRE and SynthIE-text, respectively.

\subsection{Grounding analysis}\label{subsec:IE_grounding}
In this study, we delve into the grounding efficacy of \GPTfour{}'s output.
A triplet is deemed grounded when both its subject and object entities, as well as the relation, are present in the KG.
Furthermore, for a grounded triplet to be considered correct, it must also be part of the target triplet set.

Given that being grounded is essential but not solely adequate for being correct, it is crucial to assess how well \GPTfour{}'s output aligns with the KG.
According to the data presented in \Tabref{tab:IE_grounding_analysis}, we observe that a significant portion of the output triplets from \GPTfour{} are not grounded in the KG, amounting to 45\% and 37\% on the Wiki-NRE and SynthIE-text datasets, respectively.
This finding sheds light on the importance of constrained decoding, as it ensures that the output is grounded in the KG, thereby increasing the likelihood of validity.

\begin{table}[tb]
\centering
\resizebox{0.48\textwidth}{!}{
\setlength{\tabcolsep}{7pt}
\begin{tabular}{@{}lccc|ccc@{}}
\toprule
& \multicolumn{6}{c}{Ratio of invalid triplets} \\
& \multicolumn{3}{c}{Wiki-NRE}  & \multicolumn{3}{c}{SynthIE-text} \\
& Entity   & Rel  & Triplet  & Entity   & Rel  & Triplet  \\
\midrule
\hspace{4mm} \GPTfour{}               &    \textbf{19.4}     &  \textbf{21.9}     &    \textbf{45.2}   &    \textbf{7.6}     &  \textbf{28.1}     &  \textbf{37.3}     \\
\hspace{4mm} \GPTthreeFiveTurbo{}             &    23.4     &  50.2     &    65.8   &    13.8     &  52.5     &  63.3     \\
\hspace{4mm} Claude             &   17.0     & 41.1     &    55.5   &  17.4     &  52.8     &  64.6     \\
\hspace{4mm} Claude-ins         &    19.6     &  48.4     &    62.6   &    13.8     &  43.3     &   52.7    \\
\midrule
\hspace{4mm} \ourmethod{}       &  0    &  0     &    0   &    0     &  0     &   0    \\
\bottomrule
\end{tabular}
}
\caption{\textbf{Triplets grounding analysis.}
We report the percentage of generated entities, relations, and triplets that are not present in the knowledge catalogue in few-shot unconstrained setting.
The grounding precision for \textbf{constrained} methods is 100\% by construction, and thus 0\% invalid triplets.}
\label{tab:IE_grounding_analysis}
\end{table}

\section{Task 2. constituency parsing}\label{app_sec:CP}

In this section, we provide more details about the constituency parsing task.

\subsection{Task instruction}\label{app_sec:CP_instruction}

We provide the instruction for the CP task in Figure~\ref{fig:CP_instructions}.
The few-shot demonstrations are rather long and thus we do not include them here.
The full prompt is available in our code repository.

\begin{figure}[ht]
\centering
\begin{subfigure}[b]{\columnwidth}
\centering
\begin{tikzpicture}
\node[draw, rectangle, rounded corners, fill=blue!20, inner sep=10pt, text width=0.9\columnwidth, align=center]
    at (0,0) {Perform constituency parsing on the provided sentences in accordance with the Penn TreeBank annotation guidelines. Here are a few examples.};
\end{tikzpicture}
\caption{Instruction for \sketcher{}}
\label{fig:CP_instruction_sketcher}
\end{subfigure}

\begin{subfigure}[b]{\columnwidth}
\centering
\begin{tikzpicture}
\node[draw, rectangle, rounded corners, fill=red!20, inner sep=10pt, text width=0.9\columnwidth, align=center]
    at (0,0) {In this task, you will be provided with a draft annotations that represent the parse tree of a sentence in Penn TreeBank format.  Your task is to rewrite the parse tree and fix error if any. Here are a few examples.};
\end{tikzpicture}
\caption{Instruction for \constrainedgenerator{}}
\label{fig:CP_instruction_constrained_generator}
\end{subfigure}

\caption{Instructions for parsing tasks.}
\label{fig:CP_instructions}
\end{figure}

\subsection{Constraints and grammar}\label{subsec:CP_grammar}

\begin{figure}[h!]
  \centering
    \begin{subfigure}[b]{0.4\linewidth}
    \centering
    \includegraphics[width=1.0\linewidth]{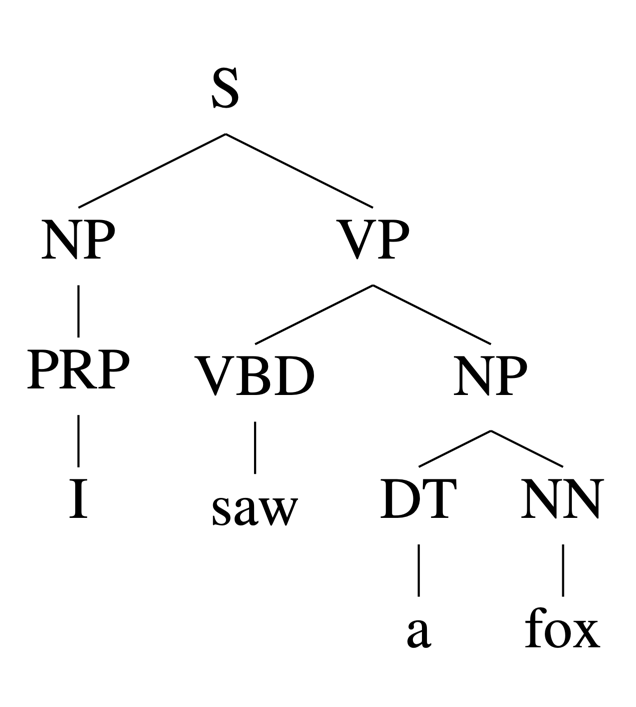}
    \caption{The correct constituency parse tree}
    \label{fig:correct_parse_tree}
    \end{subfigure}
    \hfill
    \begin{subfigure}[b]{0.4\linewidth}
    \centering
    \includegraphics[width=1.0\linewidth]{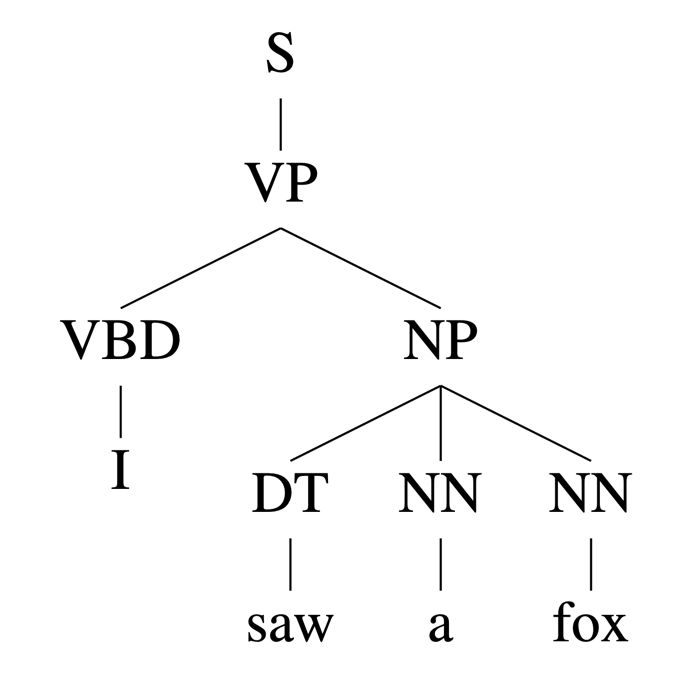}
    \caption{A grammatical but incorrect parse tree}
    \label{fig:incorrect_parse_tree}
    \end{subfigure}
    \caption{Parse trees for the sentence ``I saw a fox''.}
    \label{fig:example_parse_trees}
\end{figure}

Here we describe the grammar used to constrain the generative constituency parsing task.

\xhdr{Linearization}
A constituency parse tree is inherently a recursive structure.
To effectively represent this tree as a sequence of tokens for generation by a Large Language Model (LLM), a linearization is required.
Two common strategies for this linearization are \textit{pre-order traversal} and \textit{post-order traversal}.

We have chosen to adopt the pre-order traversal strategy.
This approach is also the default method used in the PYEVALB tool \cite{evalb} and in the construction of the Penn Treebank \cite{marcus-etal-1993-building-penn-treebank-ptb}.
As an illustration, the parse tree in \Figref{fig:correct_parse_tree} is linearized in the following format: [S [NP [PRP \textit{I}]] [VP [VBD \textit{saw}] [NP [DT \textit{a}] [NN \textit{fox}]]]].

The linearised parse tree needs to satisfy the following structural constraints:
\begin{itemize}
    \denselist
    \item \textit{Completeness}: Every word in the sentence needs to be included in the parse tree.
    \item \textit{Balanced brackets}: At any point in the linearized parse tree, the right bracket \texttt{]} should close a previously unclosed left bracket \texttt{[} and every left bracket \texttt{[} should be eventually closed by a right bracket \texttt{]}.
    \item \textit{Label consistency}: The label of terminal and non-terminal nodes needs to be consistent with the Penn Treebank format.
\end{itemize}

\xhdr{Simple Context-Free Grammar}
The tree structure of the parse tree is usually captured by a context-free grammar as shown in \Tabref{tab:cp_context_free_grammar}.

\begin{table}[h]
\centering
{\footnotesize
\begin{verbatim}
root ::= tree;
tree ::= node;
node ::= clause | phrase | word;
clause ::= spaced_open_parenthesis, space, 
            clause_tag, function_tag*, 
            index?, node*, 
           spaced_close_parenthesis;
phrase ::= spaced_open_parenthesis, space, 
            phrase_tag, function_tag*, 
            index?, node*, 
            spaced_close_parenthesis;
word ::= spaced_open_parenthesis, space, 
        word_tag, space, actual_word,
        spaced_close_parenthesis;

clause_tag ::= "S" | ...  | "SQ";
phrase_tag ::= "ADJP" | ...| "WHADVP";
word_tag ::= "CC" |...|"WRB";

function_tag ::= "-ADV" |... | "-TTL";
actual_word ::= "xxx";
index ::= "-", [1-9], {0-9};
spaced_open_parenthesis ::= space, "(";
spaced_close_parenthesis ::= space, ")";
space ::= " ";
\end{verbatim}
}
\caption{Lite Context-Free Grammar for constituency parsing.}
\label{tab:cp_context_free_grammar}
\end{table}


\xhdr{Sophisticated Regular Grammar}
However, the context-free grammar is not sufficient to capture the \textit{completeness} constraint, motivating the use of a more restrictive grammar.
\citet{geng2023grammarconstrained} proposed a sophisticated regular grammar to enforce the constraints of \textit{completeness}, \textit{balanced brackets}, and \textit{label consistency} as shown in \Figref{fig:CP_grammar}.

\begin{figure}[ht]
\centering
\begin{align*}
&S \to B_{0,0} \\
&B_{i,j} \to [\alpha \, (B_{i, j+1} \mid C_{i, j+1})]; \\
&C_{i,j} \to x_i \, (C_{i+1, j} \mid E_{i+1, j}); \\
&C_{n,j} \to E_{n, j}; \\
&E_{i, j+1} \to ]\, (E_{i, j} \mid B_{i, j}); \\
&E_{n, j+1} \to ]\, E_{n, j}; \\
&E_{n, 0} \to \varepsilon;\\
&\text{where} \, \alpha = (S \mid NP \mid VP \mid \ldots)
&\text{and} \, x_i \in \text{tokens}
\end{align*}
\caption{Sophisticated Regular Grammar for constituency parsing.}
\label{fig:CP_grammar}
\end{figure}

The grammar falls into the category of \textit{regular grammar} and is \textit{input-dependent}.
it reproduces the input sentence, represented as a sequence $x=\langle x_0, \dots, x_{n-1}\rangle$ of words, in left-to-right order, interspersing it with node labels and balanced brackets.
In order to guarantee balanced brackets, the non-terminals $B_{i,j}$ count the number of opened left brackets \texttt{[} using the second subscript index $j$, and the rules ensure that the number of closed brackets can never exceed the number of previously opened brackets.

\begin{table*}[th]
\centering
\resizebox{0.99\textwidth}{!}{
\setlength{\tabcolsep}{5pt}
\begin{tabular}{@{}lllllll@{}}
\toprule
 Method & Bracket-Prec & Bracket-Recall & Bracket-F1 & Tag Accuracy & Valid Tag & Valid Tree\\
\midrule
 \GPTfour{}     &    76.6     &    67.7  &    71.9    & 95.4 & 93.6 & 86.0\\
\hspace{4mm}  + \textbf{ Lite Context-Free Grammar}       &    \textbf{75.8}   &    \textbf{67.8}  &    \textbf{71.5}  & \textbf{95.3} & \textbf{100} & \textbf{92.5} \\
\hspace{4mm}  + \textbf{ Sophisticated Regular Grammar}       &    \textbf{69.3}   &    \textbf{63.1}  &    \textbf{66.1}  & \textbf{98.5} & \textbf{100} & \textbf{100} \\
 \GPTthreeFiveTurbo{}    &    68.2     &  55.5 &    61.2   & 93.1 & 91.7 & 76.9 \\
\hspace{4mm}  + \textbf{ Lite Context-Free Grammar}     &    \textbf{68.7}  &    \textbf{56.6}  &    \textbf{62.1}  & \textbf{92.6} & \textbf{100} & \textbf{81.5} \\
\hspace{4mm}  + \textbf{ Sophisticated Regular Grammar}      &    \textbf{61.2}  &    \textbf{49.4}  &    \textbf{54.7}  & \textbf{96.0} & \textbf{100} & \textbf{100} \\
 Claude     &    73.1     &    63.1   &    67.7 & 94.5 & 95.1 & 62.6 \\
\hspace{4mm}  + \textbf{ Lite Context-Free Grammar}     &    \textbf{71.6}  &    \textbf{62.9}  &    \textbf{66.9}  &  \textbf{93.4} & \textbf{100} & \textbf{68.7} \\
\hspace{4mm}  + \textbf{ Sophisticated Regular Grammar}     &    \textbf{52.1}  &    \textbf{45.4}  &    \textbf{48.5}  &  \textbf{75.9} & \textbf{100} & \textbf{99.2} \\
 Claude-instant     &    71.3    &    59.1 &    64.7   & 89.6 & 91.7 & 56.6 \\
\hspace{4mm}  + \textbf{ Lite Context-Free Grammar}      &    \textbf{66.6}  &    \textbf{57.4}  &    \textbf{61.6}    &  \textbf{87.9} & \textbf{100} & \textbf{67.8} \\
\hspace{4mm}  + \textbf{ Sophisticated Regular Grammar}     &    \textbf{59.6}  &    \textbf{49.2}  &    \textbf{53.9}    &  \textbf{84.9} & \textbf{100} & \textbf{99.5} \\
\bottomrule
\end{tabular}
}
\caption{\textbf{Constituency parsing with two different grammar constraints}, measured in terms of bracketing recall, precision, F1-score, and tag accuracy (with bootstrapped 95\% confidence intervals)
\dag Only subset of samples whose ground-truth parse trees are shorter than 128 tokens(\llama tokenizer) are considered, which accounts for shortest 25\% of the samples.
}

\label{tab:two_grammar_cp_results}
\end{table*}

\section{Data contamination risk}\label{sec:data-contamination-risk}

There is a rising concern over the data contamination risk of evaluating LLMs on downstream tasks.
The datasets of our experiments are publicly available on internet so there is a risk that the models may have seen the data, such as the ground true parse tree of Penn Treebank during pretraining.
However, the risk of data contamination is independent of our method and doesn't affect the validity of our conclusions.




\end{document}